\documentclass[journal,twoside,web]{ieeecolor}
\usepackage{jsen}
\usepackage{cite}
\usepackage{amsmath,amssymb,amsfonts}
\usepackage{algorithmic}
\usepackage{graphicx}
\usepackage{textcomp}
\usepackage{wrapfig}
\usepackage{bm} 
\usepackage[ruled,vlined]{algorithm2e}
\include{pythonlisting}
\usepackage{multirow}
\usepackage{caption}
\usepackage{subcaption}

\def\BibTeX{{\rm B\kern-.05em{\sc i\kern-.025em b}\kern-.08em
T\kern-.1667em\lower.7ex\hbox{E}\kern-.125emX}}
\definecolor{abstractbg}{rgb}{0.89804,0.94510,0.83137}
\begin{document}
\title{A life-long SLAM approach using adaptable local maps based on rasterized LIDAR images}
\author{Waqas Ali, Peilin Liu, Rendong Ying, and Zheng Gong
\thanks{The authors are with The School of Electronic Information and Electrical Engineering, Shanghai Jiaotong University, Shanghai, China (vaqas11@sjtu.edu.cn)}
}

\IEEEtitleabstractindextext{%
\fcolorbox{abstractbg}{abstractbg}{%
\begin{minipage}{\textwidth}%
\begin{abstract}
Most real-time autonomous robot applications require a robot to traverse through a dynamic space for a long time. In some cases, a robot needs to work in the same environment. Such applications give rise to the problem of a life-long SLAM system. Life-long SLAM presents two main challenges i.e. the tracking should not fail in a dynamic environment and the need for a robust and efficient mapping strategy. The system should update maps with new information; while also keeping track of older observations. But, mapping for a long time can require higher computational requirements. In this paper, we propose a solution to the problem of life-long SLAM. We represent the global map as a set of rasterized images of local maps along with a map management system responsible for updating local maps and keeping track of older values. We also present an efficient approach of using the bag of visual words method for loop closure detection and relocalization. We evaluate the performance of our system on the KITTI dataset and an indoor dataset. Our loop closure system reported recall and precision of above 90 percent. The computational cost of our system is much lower as compared to state-of-the-art methods. Our method reports lower computational requirements even for long-term operation. 

\end{abstract}

\end{minipage}}}

\maketitle

\section{Introduction}
\label{sec:introduction}

SLAM \cite{thrun2002probabilistic} algorithm is an integral part of an autonomous navigation system. Most applications involving autonomous robots require long-term operation in a dynamic environment. One of the main challenges for such scenarios is the problem of mapping efficiency. When a robot moves for a long time or multiple runs of the same area, the map built continues to rise and increases computational requirements. Several methods \cite{konolige}, \cite{naima2011long} have proposed efficient mapping strategies to enable long-term operation. But the problem of lifelong mapping remains unsolved especially in the case of laser SLAM. We present the method of using the rasterized images for map representation. The idea is to represent the global map as a collection of rasterized images of local maps. A map management system ensures that the online mapping remains lightweight and keeping track of all the changes in the environment. It consists of keyframes and local maps graph to track all the local maps. In addition, the system performs map updates and culling to add new information and limiting online memory usage.

For a long-term SLAM operation, the ability to re-localize and detect loop closure is also vital. If loop closure detection is not robust and the robot has to move in the same area for a long time, the SLAM system will fail. For laser SLAM interest points \cite{steder2011place, steder2010robust} or global descriptors \cite{rohling2015fast, Magnusson} has been utilized for loop closure detection. We build on the idea of using the bag of visual words for loop closure detection \cite{GalvezTRO12}. BOW has proven to be an efficient method for content-based image retrieval from a large database. Several vision-based SLAM algorithms \cite{KEJRIWAL201655, angeli2008real, nicosevici2012automatic} use the bag of words based methods for loop closure detection. In this paper, we propose the BOW approach for the loop closure method using rasterized laser images. One of the drawbacks of the BOW approach is the computational complexity of building vocabulary. We represent the global map as a set of local maps in the form of rasterized images. So only a limited number of local map images are used to build the BOW database making the system more efficient. We have used KITTI and an indoor dataset to evaluate the performance of our system. We use several state-of-the-art methods for comparison to show the efficiency of the proposed approach. Our system reported a higher recall and precision rate with much lower computational requirements. We present the following contributions in this paper.

\begin{enumerate}
	\item A novel mapping strategy for laser-based SLAM system. We propose the representation of the global map as a collection of rasterized images built from local maps.
	
	\item A map management system based on a graph structure of local maps and keyframes poses with an efficient update and culling algorithm. 
	
	\item We propose a novel loop closure detection and re-localization for laser SLAM system based on the BOW approach.
	
	\item A lightweight SLAM algorithm with much lower computational requirements for long-term application
\end{enumerate}

\section{Related Work}

One of the important aspects of the laser SLAM system for long-term operation is an efficient and robust loop closure and re-localization method. Hess et al. \cite{Hess} presented a method of loop closure detection using scan-to-submap search. Mapping is divided into sub-maps and as each submap is finished it is qualified for loop closure search. Appearance-based loop closure detection is proposed in \cite{Magnusson}. Features are detected based on NDT surface representation and by matching these features loop closure is detected. Steder \cite{steder2011place} proposed an approach of applying the bag of words method for place recognition using laser range images. Their method reported a good recall rate but its time requirements for interest point detection, validation, and histogram calculation is high. We propose an efficient implementation of BOW loop closure using a small online built vocabulary based on local maps. Behley and Stachniss \cite{Behley} proposed a SLAM system using surfel based maps and loop closure is detected by matching the new scan with the rendered map. 

In our earlier work \cite{ali20216}, loop closure detection is performed in two steps i.e. first finding closest keyframes and then using feature matching to confirm loop closure candidate. In this paper, we apply the bag of words approach for loop detection. We make this method more efficient by searching through a database built from local map images. Our method reported high precision and recall rates in real-time.

Most real-time robot applications require long term operation in a dynamic environment. The earliest solution to navigation in dynamic environments was presented by Yamauchi et al. \cite{yamauchi}. They proposed reactive behavior and adaptive place network to deal with dynamic scenes, but their drawback is that it cannot self localize and provides no solution to exploration in unknown scenes. Stachniss and Burgard \cite{stachniss2005} presented a method for modeling configuration of the semi-static environment. Their method clusters local maps to give an estimate of the possible configuration of the environment. Bosse et al. \cite{bosse2004} presented the Atlas framework for realizing large scale real-time operation using sub-maps. It uses a collection of sub-maps with their local frames, and a graph structure is formed based on vertices consisting of sub-maps and edges as a transformation between these local maps.

In \cite{biber2005dynamic,biber2009experimental} a dynamic representation of map is presented which consists of local maps built at different times, these maps are continuously updated online. Those maps are selected for navigation which best suits the current observation. This operation requires a large amount of memory to update the maps online. Our method uses a rasterized image representation for 3D laser data and the map updates require a much lesser computational cost. In \cite{kretzschmar2011efficient, stachniss2017pose} a method for limiting the computational complexity for long term operation using information-theoretic compression of pose graph is presented. Observations that provide little information are discarded and in this way nodes from pose-graph are marginalized. New nodes are only added when the robot explores previously unknown scenes. Our method provides an efficient way to store the maps while also keeping older information for long-term operation. 

McDonald et al. \cite{McDonald} proposed a system for multi-session mapping. It incorporates multiple maps using a single coordinate and appearance-based loop closure detection is used. Fentanes et al. \cite{fentanes2015now} proposed a method to predict the changes in an environment based on periodic events this allows for long term robot operation. Wolf and Sukhatme \cite{wolf2005mobile} proposed a mapping method for dealing with static and dynamic object separately. These environment models limit the application of the system to certain scenarios, while we build a system that can successfully work in different conditions and is not limited to specific environments. Konolige and Bowman \cite{konolige} proposed a method for mapping in dynamic environments for long term operation. A criterion for system deployment and practical life long mapping is defined in their work. Robust visual place recognition is used for loop closure detection. It can recover maps that may have been distorted due to localization failure and stitch different sequences. Banerjee et al. \cite{banerjee} proposed a method to limit computational cost while keeping efficiency and consistency by pruning redundant local maps. Our method is both computationally efficient and able to relate changes in a dynamic environment. We save the local maps in the form of rasterized images, which require very little memory and life-long mapping is done online.

\begin{figure}[htp]
	\centering
	\includegraphics[scale=0.37]{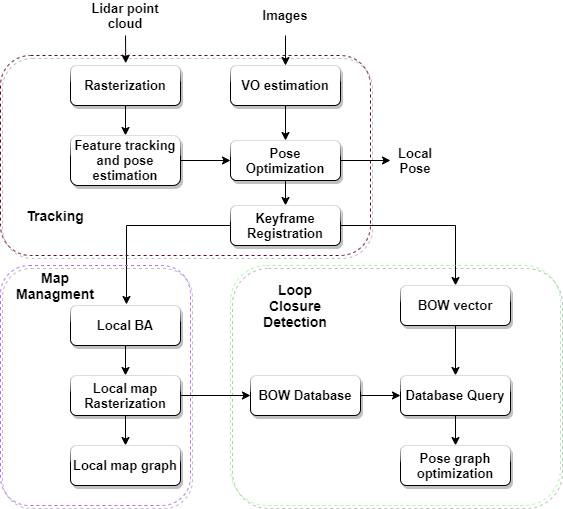}
	\caption{Complete architecture of our SLAM system}
	\label{fig:sys}
\end{figure}

\section{Tracking}

\subsection{Visual Odometry}

The purpose of including the visual odometry into our system is to ensure that the tracking does not fail in unstructured or feature-less environments. There are several state-of-the-art visual odometry methods present in the literature. In this paper, we use the approach presented in \cite{mur2017orb}. There are two main parts of odometry calculation, i.e., pose estimation from feature tracking and local BA. VO thread starts by taking in the images and extract ORB features \cite{rublee2011orb}. ORB feature detector is stable, fast to compute, and works in a variety of environments. We set the scale level for FAST corners at 8, and each scale is further divided into grid cells to get even distribution of features. We ensure that enough features are extracted from each image by adapting the threshold parameters of the FAST corner detector. Using a constant motion model initial pose is estimated. Next, we perform a search for map points. The pose estimate is first optimized from the map-point correspondences and then using motion BA. The final pose is then used for fusion with the laser pose.

\subsection{Laser Pose Estimation}

\begin{figure}
	\centering
	\includegraphics[scale=0.2]{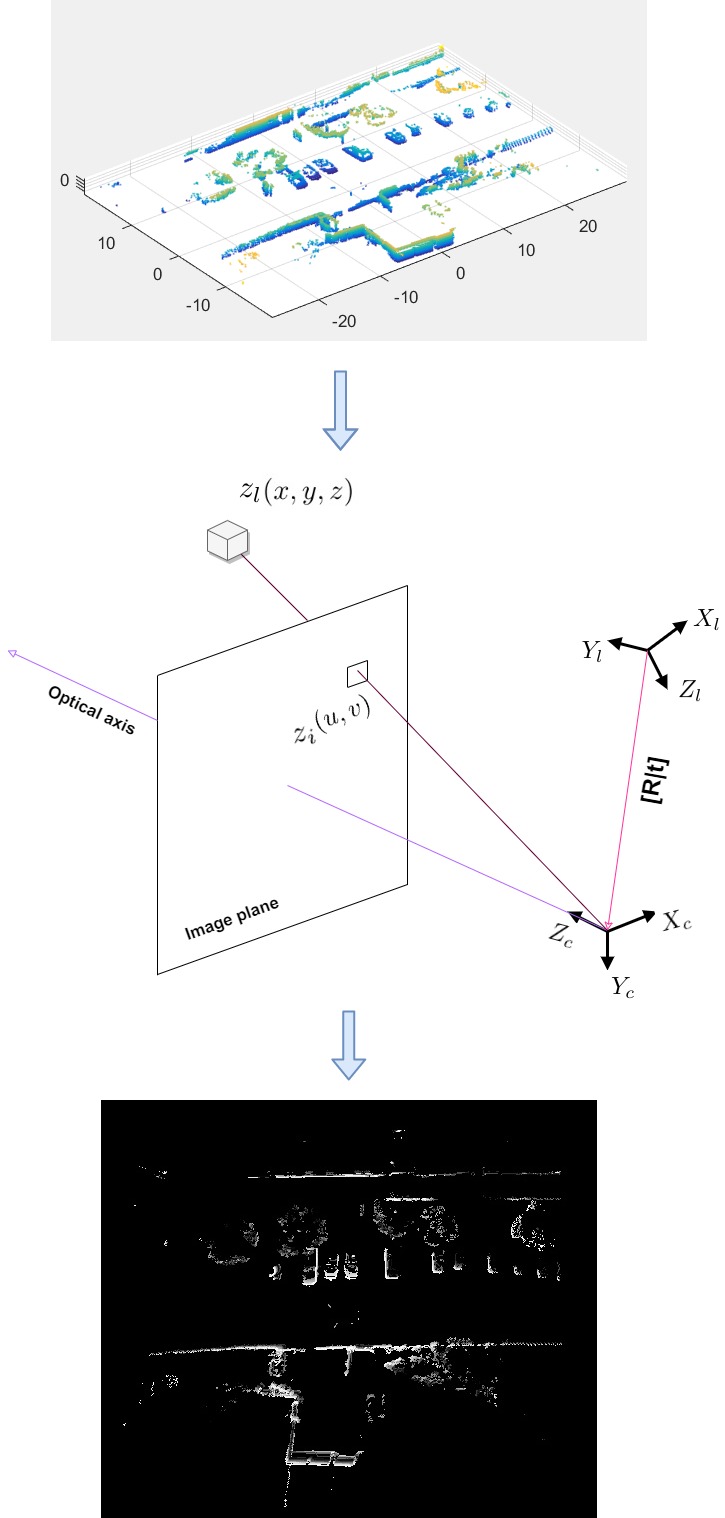}
	\caption{The raw point cloud is first processed to remove outliers and then rasterized to form a greyscale image.}
	\label{fig:track}
\end{figure}

Pose estimation from the laser point cloud is based on our earlier work \cite{ali20216}. There are three main components of this thread, i.e., point cloud rasterization, feature extraction and matching, and pose estimation. Once the raw point cloud is received, it is processed to remove outliers. For the 3d point cloud collected by multi-line lidar, circular rings are formed on the ground plane. Such rings can cause outliers when projected to the image. So we apply RANSAC \cite{fischler1981random} based plane detection to remove the ground plane from the raw 3D point cloud. The remaining point cloud is used for rasterization. We use a pinhole camera model to give a geometric relationship between 3d points and their 2d projections. Figure \ref{fig:track} illustrates the process of rasterization. 

The 3d points received are in lidar coordinates, which are first transformed to camera coordinates using extrinsic parameters defined by a rotation matrix $\textbf{R}$ and a translation vector $\textbf{t}$. Next, we use the intrinsic parameters $\textbf{K}$ to project points to image coordinates. The final relationship for projecting a 3d point $z_l$ in lidar coordinates to 2d pixel $z_i$ in image coordinates is given in equation \ref{eq.pinh}. We set the z-axis as the optical axis during the process of rasterization. The pixel intensity of each point on the image is set equal to the $z$ value of its corresponding 3d point. In this way, we get a greyscale image, and the elevation information of 3d points is saved in the image.

\begin{equation}
	z_i = \bm{K}\bm{R} (\bm{I} | \bm{t}) z_l 
	\label{eq.pinh}
\end{equation}

Next, we apply the ORB feature \cite{rublee2011orb} detector on the rasterized images. At least 1000 features are detected from each image. We match features with the last scan and remove outliers using RANSAC \cite{fischler1981random}. Once feature correspondences are found, these features are projected back to laser coordinates. For a set of feature points $P = \{p_1,..., p_N\}$ and $Q = \{q_1,...,q_N\}$ from two scans. We use eq.\ref{eq.icp} to estimate the transformation in the form of rotation matrix $\textbf{R}_l$ and translation vector $\textbf{t}_l$ between the two scans.

\begin{equation}
	f\left(\bm{R}_l,\bm{t}_l\right) = \frac{1}{N_p}\sum_{i=1}^{N_p} ||q_i - \bm{R}_l p_i - \bm{t}_l||^2
	\label{eq.icp}
\end{equation}

We use the ICP algorithm to estimate motion from these points. The number of points is low with known correspondences, therefore ICP gives an accurate estimate. We use the motion estimate to calculate the lidar pose. The tracking thread is also responsible for keyframe selection. We use simple criteria of keyframe selection, i.e., at least five frames have passed and there are at least 100 points in common with the last frame. Finally, we build a factor graph based on keyframe nodes, where each node represents the estimated pose from the laser point cloud. We then add visual odometry factors into the factor graph and optimize to calculated the final keyframe pose. Each keyframe stores the 6dof pose and 3d values of all the features observed at that position. The keyframes information is used by mapping and loop closure threads.

\subsection{Re-localization}

One of the main contributions of this paper is the design of the relocalization approach for laser-based SLAM. It is also vital for the long-term performance of robot localization. The principle for relocalization is similar to loop closure detection. In the event, enough features are not detected from laser rasterized images and VO doesn't provide an accurate pose estimate, meaning that the tracking is lost. We convert the laser rasterized image from the current frame into a BOW vector and perform a database query. After getting a match from the database, we calculate the robot pose by matching the features from the current frame with the frame received from a database query. The process of building the BOW database from local maps is discussed in detail in the next section.

\begin{figure}[htbp!]
	\centering
	\includegraphics[width=3in]{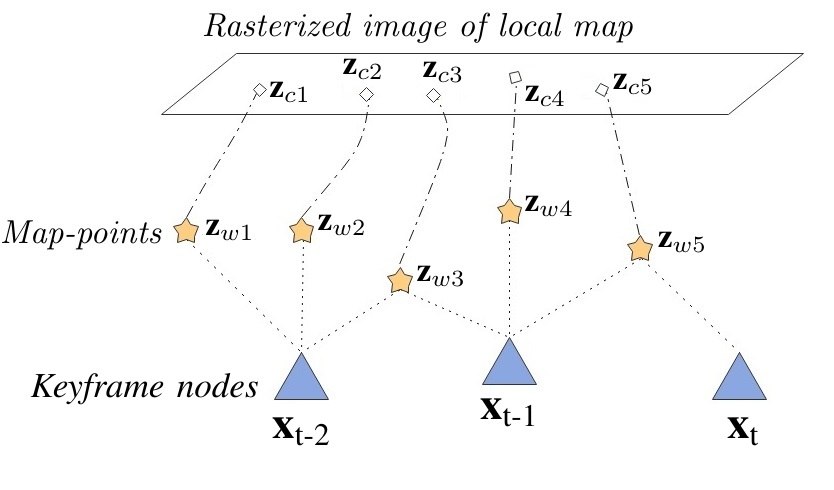}
	\caption{Mapping process of rasterizing the points of local map to form rasterized image}
	\label{fig:map}
\end{figure}

\section{Mapping}

The mapping thread takes the keyframe information from Tracking to register the local map and perform local BA. It starts by taking the map points observed at the first keyframe to initialize the local map. The local map is built incrementally as the information of the new keyframe is received. New map points are added to the local map until it meets the size threshold. At that point, a new local map is initialized. We ensure a smooth transition by keeping enough common points between consecutive local maps. When a local map is finished, we perform local BA to optimize the map points and keyframe poses. The cost function to solve local BA is defined as

\begin{equation}
e_{ij} = \textbf{z}_{ij} - \hat{\textbf{z}}_{ij}(\textbf{x}_{i},\textbf{x}_{j})
\label{eq.cost}
\end{equation} 

In equation-\ref{eq.cost}, $\textbf{x}_{i}$ and $\textbf{x}_{j}$ denote the poses for keyframe i \& j and $\hat{\textbf{z}}_{ij}$ is the predicted observation using $\textbf{x}_{i}$ and $\textbf{x}_{j}$. The actual observation is $\textbf{z}_{ij}$, giving the cost function $e_{ij}$ equal to the difference of $\textbf{z}_{ij}$ and $\hat{\textbf{z}}_{ij}$. We minimize the re-projection error by solving eq. \ref{eq.func}.

\begin{equation}
\textbf{F}(x^*) = argmin \sum_{ij}e^T_{ij}\bm{\Omega}_{ij}e_{ij}
\label{eq.func}
\end{equation}  

After the local BA is finished, the optimized map points within the local map are rasterized to form an image using eq. \ref{eq.pinh}. Fig-\ref{fig:map} shows the mapping process, $\{\textbf{z}_{w1},...,\textbf{z}_{w5}\}$ are optimized 3d map points of a local map. These points are rasterized to form the local map image. In this way, the global map is represented as a collection of local map images. Once a local map is rasterized to an image, it is forwarded to the BOW database and map management system.

\begin{figure}[htbp!]
	\centering
	\includegraphics[scale=0.4]{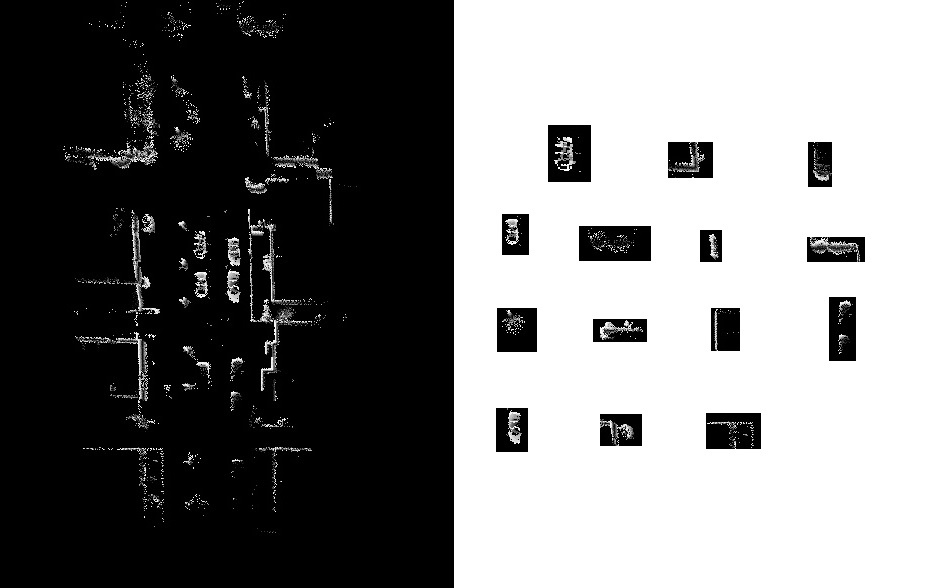}
	\caption{A set of visual word extracted from a rasterized image of local map}
	\label{fig:bow}
\end{figure}

\subsection{BOW Database}

The mapping thread is also responsible for maintaining the BOW database. The rasterized local map images are used to build a database based on the bag of words approach \cite{GalvezTRO12}. BOW is most commonly used for vision-based SLAM algorithms to detect loop closure. Here we adopt this approach to apply for laser-based rasterized images. Instead of building the database of all the keyframes, we only use local maps for the database. This reduces the dimensions of the database and makes the whole system more efficient.

For every local map image, we extract ORB features similar to the tracking thread. We compute FAST corners and assign BRIEF descriptors to each corner. The next step is to build a vocabulary using DBOW2 \cite{GalvezTRO12} based on these features. The DBOW2 library discretizes binary descriptors, applies k-median clustering, and builds a vocabulary tree of these clusters. Figure-\ref{fig:bow} shows the example of visual words extracted from a rasterized local map image. We see three types of features dominating the visual words, i.e., building structures such as corners or planes, trees or shrubs, and dynamic objects such as cars. Using our approach, we can save the structural information of the scene. So even if there are some changes in the scene, we can still detect loop closure efficiently.

\begin{algorithm}
	\SetAlgoLined
	\KwIn{Keyframe $\textbf{x}_i$, Features $f_i \Rightarrow \{z_{i1},..., z_{in}\}$}
	\KwResult{Perform Mapping for long-term SLAM operation }
	initialization\; 
	\While{Recieve Keyframes}{
		initialize local map $\mathcal{L}$\;
		\eIf{local map finished}{
			perform local BA\;
			rasterize map-points to local map image $I_t$\;
			add time and keyframe label to $I_t$\;
			\eIf{local map graph present}{
				initialize local map graph\;
			}{
				add $I_t$ to graph\;
			}
			\If{loop closure detected}{
				initialize local map graph update\;
				perform culling\;		
			}
			\If{distance travelled greater than $D_{th}$}{
				perform culling\;
			}
		}{
			add Keyframes to local map\;
		}
	}
	\caption{Life-long Mapping}
\end{algorithm}

\subsection{Map Management}

\subsubsection{Initialization}
This first step for our map management scheme is to initialize a graph structure based on the local map images and their relation to the keyframes. We know the covisibility information of the landmarks $\textbf{z}_w$ concerning each keyframe $\textbf{x}$. The relation of points $\textbf{z}_w$ to a pixel point $\textbf{z}_c$ is based on equation.\eqref{eq.pinh}. Figure-\ref{fig:map} shows the relationship between keyframes, map points, and local map images. Our idea here is to use these relations and build a graph that only includes keyframes' poses and local map images. In long-term operation, we can use this graph to track all the maps. It is also useful to perform an update and remove redundant information. When initializing a local map into our graph each map is tagged with time information, which becomes useful later when the robot revisits an area. So in the graph structure, each local map has two labels, i.e., the keyframes indices to which the local map is connected and the time stamp.

While an online local map graph is kept, we also keep an offline database that stores the older local maps. So the older information is not completely lost but saved into the offline database.

\subsubsection{Map Update}

The updates to the local maps are vital for the long-term operation of our system. When a robot revisits an area or loop closure is detected, we want to update the information fused in local maps. When performing an update to the local map graph, the goal is not to lose older observations. When loop closure is detected a new local map is initialized at that position using the current information provided by laser scans. It is tagged with the time information and we remove older maps at the position from the online graph to save them into the offline database. 

\subsubsection{Culling}

To keep the mapping efficient, we design a robust culling scheme for local maps. The culling mechanism has two parts, i.e. when the loop is detected and based on distance traveled. As discussed in the last section that when the loop is detected we add a new local map to the graph. The first part of the culling method is applied here to remove the older local map from the graph. For the second part, we set a distance threshold. If the traveled distance of the robot meets the threshold then older local maps are removed from the online graph. The final goal of the system is to minimize the online information while also maintaining system accuracy. We save all the marginalized local maps into an offline database.

The culling of older maps does not affect the global operation because we perform the re-localization and loop closure detection based on frame query from the BOW database. Using the proposed strategy the online operation remains efficient and robust to changes in the environment.

\section{Loop Closure}

\begin{figure}[htbp!]
	\centering
	\includegraphics[scale=0.3]{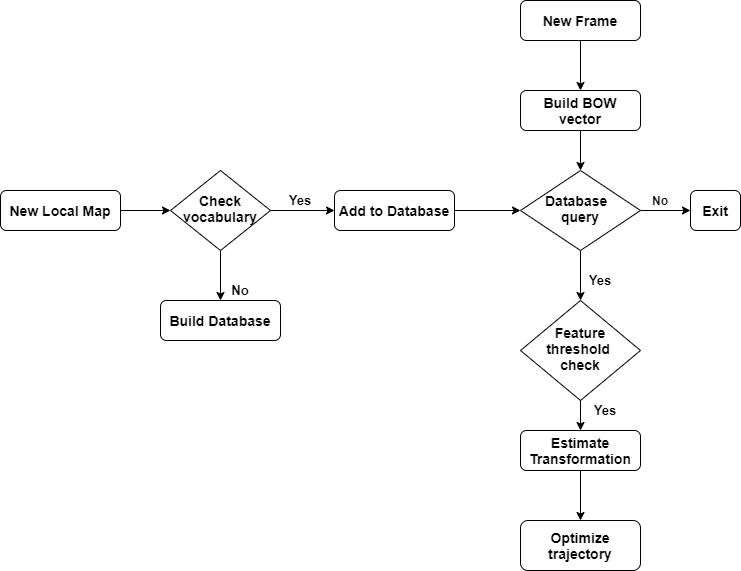}
	\caption{Control flow for loop closure detection where each new frame is used for database query from BOW database}
	\label{loop_fl}
\end{figure}

Loop closure detection for the system works in three steps i.e. database query, feature matching, and transformation estimation. Figure-\ref{loop_fl} shows the algorithm flow for loop closure detection. A BOW database is managed by the mapping thread based on local maps. Our method has the advantage that minimizes the computational time required for loop candidate detection as the image space is reduced by just searching through the local maps. The method starts by taking the descriptors of a new frame $F_i$ and converting it to a BOW vector and performing a database query. We use the DBOW2 \cite{GalvezTRO12} library to perform the database query. A vector of the matches found from the query with normalized scores of each match is returned.

From the list of matches, we apply a score threshold. We select the matches of the current keyframe with local maps that pass the threshold as initial candidates. Then, we match features from the current keyframe to each keyframe within the local map. Here the local map graph becomes useful as each local map contains a keyframes label. We use the RANSAC algorithm to detect outliers in the feature matches. If the current keyframe $\textbf{x}_i$ reports enough correspondences with keyframe $\textbf{x}_j$, then this candidate can be considered as the final loop closure. For global optimization, we build a pose-graph, where landmarks are marginalized. We need to estimate a 6DOF transformation between frame $\textbf{x}_i$ and $\textbf{x}_j$, for adding loop constraint in the pose graph. Using the already estimated correspondences between the two frames, we compute the transformation matrix. Finally, we optimize the trajectory after adding the loop closure constraint. 

\section{Experiments}

We evaluate our method using KITTI \cite{Geiger2012CVPR} odometry dataset and an indoor dataset. KITTI dataset provides 22 sequences with 3D lidar data and stereo images. Sequences 00 to 10 serve as training datasets as ground truth data is provided. For the KITIT dataset, Velodyne HDL-64 LIDAR is used to record laser data. They record the data in three types of environments, i.e., urban, country, and highway. In addition, we record an indoor dataset with a VLP-16 laser scanner. This indoor dataset is recorded with multiple runs through our lab. It is useful to show the long-term SLAM performance in a dynamic environment.

We evaluate in two steps, i.e., first, we use the KITTI dataset to show the performance of the SLAM system in terms of localization accuracy, loop closure detection, and computational requirements. Next, we use the indoor dataset to evaluate the performance of our system in a long-term operation.

\subsection{KITTI dataset}
\subsubsection{Localization accuracy}

The localization performance of the system is assessed using relative translation and rotation errors based on the method presented in \cite{Geiger2012CVPR}. The relative rotation and translation errors are computed using equations \ref{eq_rot} and \ref{eq_trans}.

\begin{equation}
E_{rot} (\mathcal{F}) = \frac{1}{|\mathcal{F}|}\sum_{(i,j)\in \mathcal{F}}\angle[(\hat{p_j}\ominus\hat{p_i})\ominus(\hat{p_j}\ominus\hat{p_i})]
\label{eq_rot}
\end{equation}
\begin{equation}
E_{trans} (\mathcal{F}) = \frac{1}{|\mathcal{F}|}\sum_{(i,j)\in \mathcal{F}}\parallel(\hat{p_j}\ominus\hat{p_i})\ominus(\hat{p_j}\ominus\hat{p_i})\parallel_2
\label{eq_trans}
\end{equation}

In equations \ref{eq_rot} \& \ref{eq_trans}, $\mathcal{F}$ is a set of frames $(i,j)$, $p$ and $\hat{p}$ are $SE(3)$ ground truth and estimated poses respectively, $\angle[.]$ denotes rotation angle and $\ominus$ is inverse compositional operator.

The localization performance is compared with LOAM \cite{loam}, GICP \cite{segal} and VINS \cite{qin2019a}. LOAM is the best-performing laser-based method on the KITTI dataset and VINS is one of the best performing visual odometry methods. We run the open-source code available for the three systems on the eleven training sequences provided by the KITTI dataset. Table-\ref{tab:accuracy} shows the values of relative translation and rotation errors for the 11 sequences. Our system consistently gives accurate performance for all the training sequences for both translation and rotation errors, while also surpassing other methods in some sequences. The values of our method remain low for almost all the sequences, while the results of remaining three methods vary for the 11 sequences. We can prove that our method can produce accurate results on par with state-of-the-art methods.

\begin{table*}[hbt!]
	\centering
	\begin{tabular}{ccccccccc}
		\hline
		\multirow{2}{*}{Sequences} & \multicolumn{2}{c}{Our Method}    & \multicolumn{2}{c}{LOAM}        & \multicolumn{2}{c}{VINS} & \multicolumn{2}{c}{GICP}      \\
		& Trans           & Rot             & Trans         & Rot             & Trans       & Rot        & Trans         & Rot           \\
		\hline
		0                          & \textbf{1.2275} & 0.0061          & 1.1           & \textbf{0.0053} & 1.3517      & 0.012      & 1.29          & 0.64          \\
		1                          & \textbf{1.4493} & \textbf{0.0032} & 2.79          & 0.0055          & 2.2273      & 0.0076     & 4.39          & 0.91          \\
		2                          & \textbf{1.0808} & \textbf{0.0052} & 1.54          & 0.0055          & 1.3989      & 0.009      & 2.53          & 0.77          \\
		3                          & \textbf{0.8401} & \textbf{0.0034} & 1.13          & 0.0065          & 1.2116      & 0.0108     & 1.68          & 1.08          \\
		4                          & \textbf{0.5469} & \textbf{0.0044} & 1.45          & 0.005           & 1.4229      & 0.0118     & 3.76          & 1.07          \\
		5                          & 1.0494          & 0.0045          & \textbf{0.75} & \textbf{0.0038} & 1.4272      & 0.012      & 1.02          & 0.54          \\
		6                          & 1.2673          & 0.0042          & \textbf{0.72} & \textbf{0.0039} & 1.2926      & 0.011      & 0.92          & 0.46          \\
		7                          & 1.4091          & 0.0078          & 0.69          & 0.005           & 1.3308      & 0.013      & \textbf{0.64} & \textbf{0.45} \\
		8                          & \textbf{1.0798} & \textbf{0.0053} & 1.18          & 0.0044          & 1.8352      & 0.012      & 1.58          & 0.75          \\
		9                          & 1.495           & 0.005           & \textbf{1.2}  & \textbf{0.0048} & 1.6345      & 0.0085     & 1.97          & 0.77          \\
		10                         & \textbf{0.8396} & \textbf{0.0053} & 1.51          & 0.0057          & 3.1007      & 0.0185     & 1.31          & 0.62         \\ \hline
	\end{tabular}
	
	\caption{Relative translation and rotation error for our method, LOAM, VINS and GICP}
	\label{tab:accuracy} 
\end{table*} 

\subsubsection{Loop Closure}

In this section, we evaluate the performance of the loop closure detection method presented in this paper. It is one of the main contributions of this paper. For evaluation, we have implemented two-loop closure techniques i.e. by searching through keyframes \cite{ali20216} and searching through local maps. In these two methods, the search for loop closure is performed based on feature matching. In the first method, we search for the nearest keyframes based on distance threshold and then match features to give the final loop candidate. The second method is similar but uses search through local maps instead of keyframes.

\begin{table*}[hbt!]
	\centering
	\begin{tabular}{ccccccc}
		\hline
		\multirow{2}{*}{Sequence} & \multirow{2}{*}{\begin{tabular}[c]{@{}c@{}}Total \\ frames\end{tabular}} & \multirow{2}{*}{\begin{tabular}[c]{@{}c@{}}No. of \\ Keyframes\end{tabular}} & \multirow{2}{*}{\begin{tabular}[c]{@{}c@{}}No. of \\ Local maps\end{tabular}} & \multicolumn{3}{c}{Time for candidate detection (msec)} \\
		&  &  &  & BOW & \begin{tabular}[c]{@{}c@{}}Search through \\ Local maps\end{tabular} & \begin{tabular}[c]{@{}c@{}}Search through \\ Keyframes\end{tabular} \\ \hline
		00 & 4540 & 1125 & 225 & 3.034 & 130.49 & 537.1 \\
		02 & 4660 & 1185 & 237 & 6.64 & 402 & 631 \\
		05 & 2760 & 870 & 174 & 2.1 & 103.70 & 384.60 \\
		06 & 1100 & 280 & 56 & 1.29 & 52.83 & 28.90 \\
		07 & 1100 & 265 & 53 & 1.102 & 21 & 26 \\
		09 & 1590 & 395 & 79 & 1.338 & 23 & 12.75 \\ \hline
	\end{tabular}
	\caption{Detailed values for each sequence used in loop closure evaluation}
	\label{tab:loop_time}
\end{table*}

\begin{figure}[!h]
	\centering
	\includegraphics[scale=0.35]{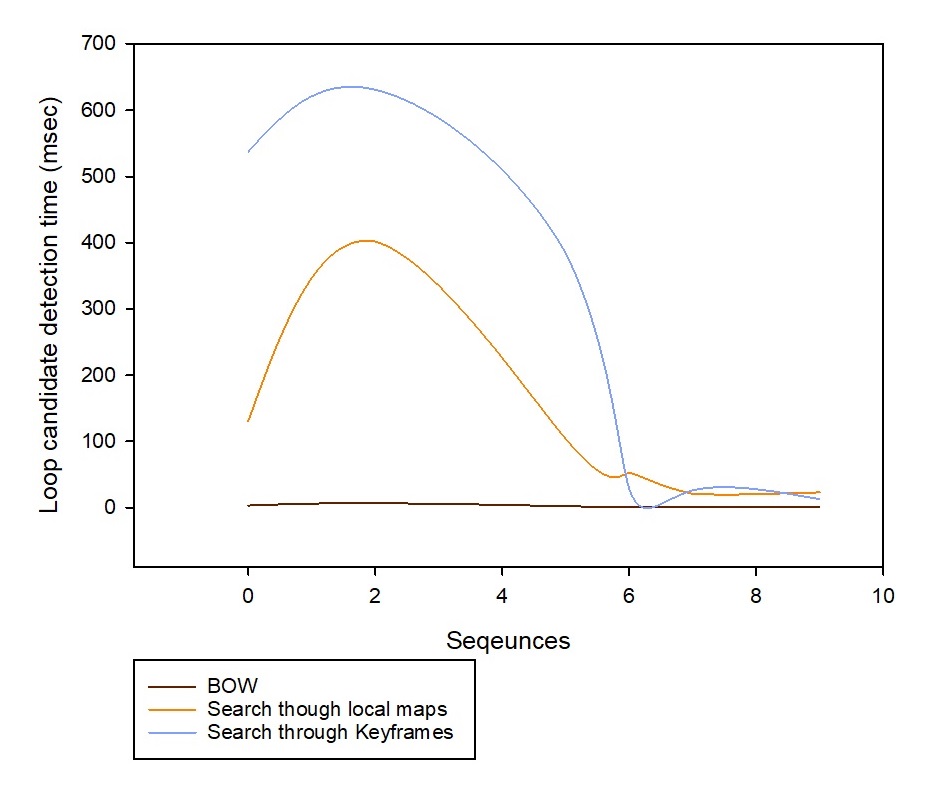}
	\caption{Loop candidate detection time comparison for BOW, local map and keyframe search loop detection methods}
	\label{fig:loop_stat}
\end{figure}

We use six sequences with loop closure from the training sequences of the KITTI dataset. Table-\ref{tab:loop_time} shows the stats for the six sequences, i.e., the total number of frames, and the number of keyframes and local maps selected by our system. The average time taken by each method for loop closure candidate detection is also given in the table. The approach of using local maps greatly reduces the space from which we can search for loop closure. Take the example of sequences 00 and 02, the longest sequences with loop closures. It contains 4540 and 4660 frames, which can be represented by 225 and 237 local maps respectively. So we get an efficient representation of the complete map. Next, we have a look at the time taken by each method for loop candidate detection. Firgure-\ref{fig:loop_stat} shows the plot for loop detection time requirements of each method.  

\begin{table*}[!h]
	\centering
	\begin{tabular}{ccccccc}
		\hline
		\multirow{2}{*}{Sequence} & \multicolumn{3}{c}{Recall (\%)} & \multicolumn{3}{c}{Precision (\%)} \\
		& BOW & \begin{tabular}[c]{@{}c@{}}Search through \\ Local maps\end{tabular} & \begin{tabular}[c]{@{}c@{}}Search through \\ Keyframes\end{tabular} & BOW & \begin{tabular}[c]{@{}c@{}}Search through \\ Local maps\end{tabular} & \begin{tabular}[c]{@{}c@{}}Search through \\ Keyframes\end{tabular} \\ \hline
		00 & 94.78 & 98.2 & 72.25 & 98.5 & 100 & 100 \\
		02 & 84.28 & 89.28 & 59.2 & 80.5 & 94 & 78.13 \\
		05 & 88.58 & 97.59 & 80.18 & 92.8 & 100 & 100 \\
		06 & 84.5 & 98.89 & 51 & 100 & 100 & 100 \\
		07 & 69.23 & 84.6 & 61.53 & 100 & 100 & 100 \\
		09 & 65.21 & 82.6 & 39.13 & 100 & 100 & 100 \\ \hline
	\end{tabular}
	\caption{Values of recall and precision for loop closure detection using BOW, local map search and keyframe search}
	\label{tab:rcl_pr}
\end{table*}

\begin{figure}
	\centering
	\includegraphics[scale=0.35]{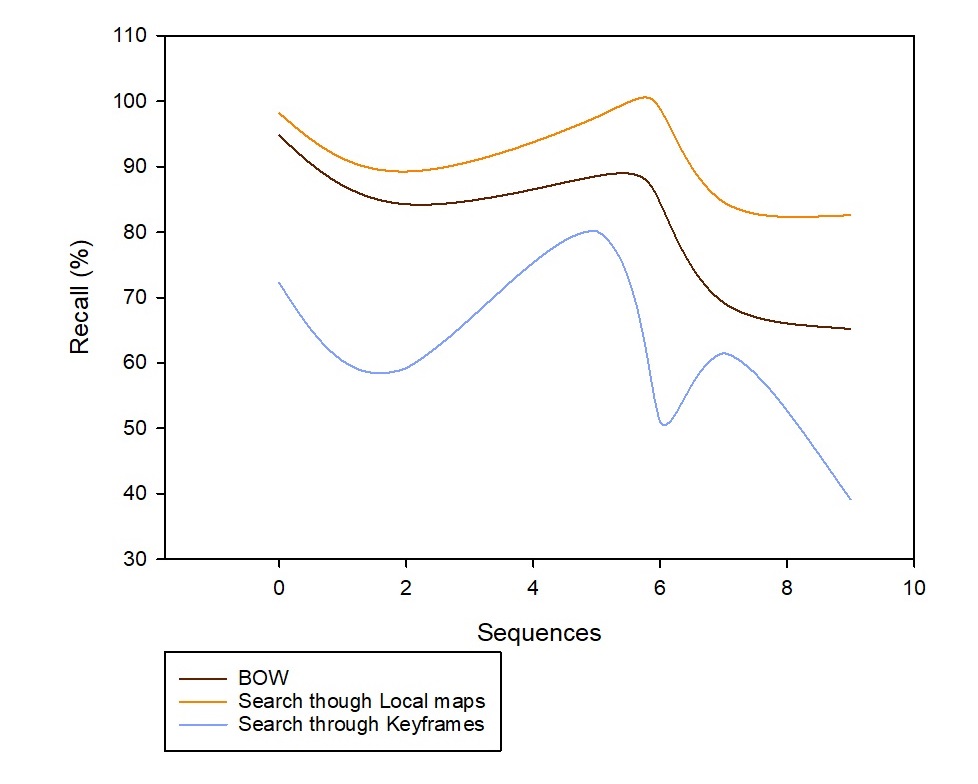}
	\caption{A comparison of loop closure detection recall rate for the three methods}
	\label{fig:recal}
\end{figure}

The method presented in this paper has a major difference in performance. The average time taken by searching through local maps and keyframes is 130.49ms and 537.1ms respectively, while the time required by BOW based method takes only 3.034ms. Bag of words has been proven to be a very efficient approach that has been widely used in vision-based systems, but we adapt this method to apply to laser-based SLAM system. We make this approach time-efficient by building the database based on local maps. Table-\ref{tab:rcl_pr} shows the values for recall and precision for the three loop detection methods and a plot is given in figure-\ref{fig:recal}. All the methods present high precision, but the keyframe-based method has a slightly lower recall rate because it is slower and can miss some frames. BOW method has a slightly lower recall rate as compared to feature-based search through local maps, but keeps high precision. These results validate the efficacy of the method presented in this paper. We showed that BOW-based loop closure detection for laser SLAM gives a precision of more than 90\% and recall rate of above 80\% and only needs few milliseconds to detect loop candidate.

\subsubsection{Computational Complexity}

In this paper, one of our main contributions is to build a lightweight SLAM algorithm. We select the six longest sequences from the KITTI odometry dataset, i.e., 00, 02, 08, 13, 19, and 21 to evaluate the computational requirements of our system. Out of these sequences 00, 02, 13, and 19 contain loop closures. This also gives us an idea of the computational requirements during loop closure detection. We have selected five SLAM methods for evaluation in this section including ORBSLAM2 \cite{mur2017orb}, Cartographer \cite{Hess}, ISCLOAM \cite{iscloam}, MULLS \cite{mulls} and SUMA++ \cite{sumapp}. All of these methods contain loop closure detection. ORBSLAM2 is included in our evaluation because it also uses DBOW2 \cite{GalvezTRO12} to detect loop closure and relocalization. We want to compare the performance of our system with both laser and vision-based systems. Further, the cartographer requires imu data and the KITTI dataset only provides imu data for sequences 0 to 10. So we test cartographers only on sequences 00, 02, and 08.

\begin{figure}[hbt!]
	\centering
	\includegraphics[scale=0.55]{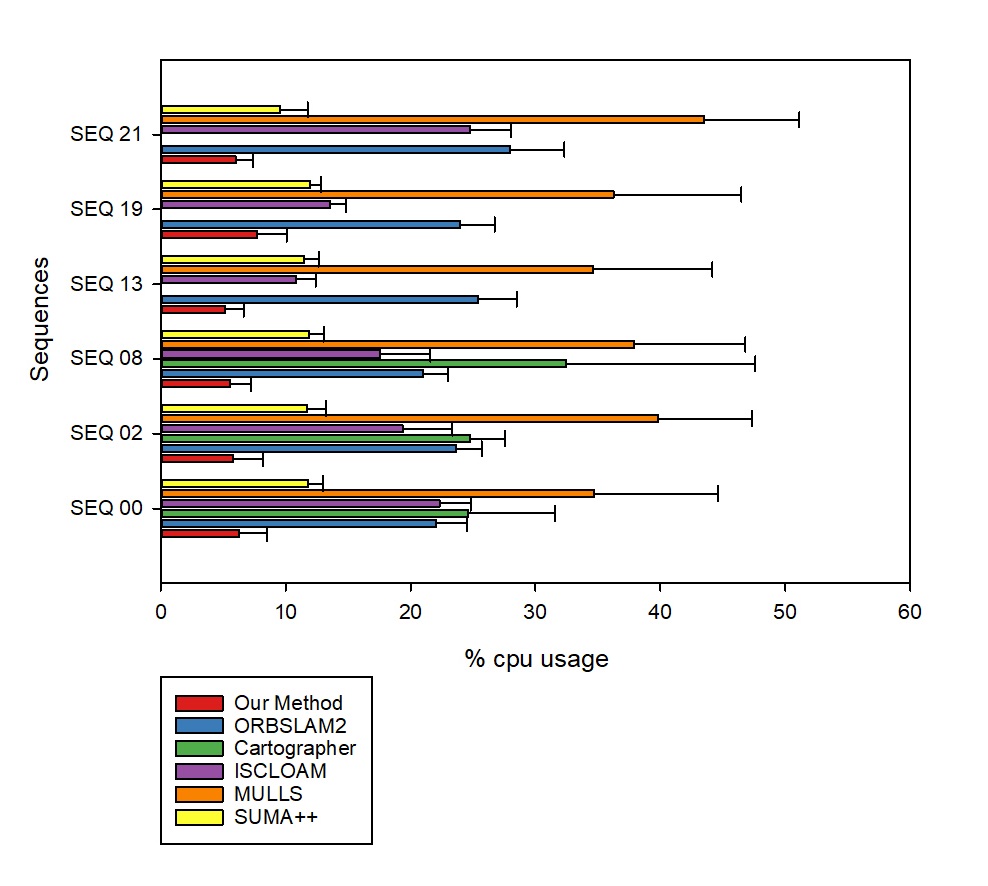}
	\caption{Results of the cpu usage of our method in comparison with ORBSLAM2, cartograper, ISCLOAM, MULLS and SUMA++ for KITTI dataset}
	\label{fig:cpu_us}
\end{figure}

We select two metrics to measure the computational cost for each method, i.e., the percentage CPU and memory usage. First, we look at the CPU usage, figure-\ref{fig:cpu_us} shows the values for all the six methods. The requirements of our system consistently remain low for all sequences, as the values remain less than 10\%. The reason is that we propose a simple approach for the SLAM. The pose estimation is done by feature extraction and matching, then mapping performs only local BA and saves maps as rasterized images. Finally, loop closure detection is performed by search through a small BOW database. Looking at other systems, the CPU usage is much higher for all methods, except SUMA++. It has slightly lower requirements as it also uses laser image representation. 

\begin{figure}[hbt!]

	\centering
	\includegraphics[scale=0.55]{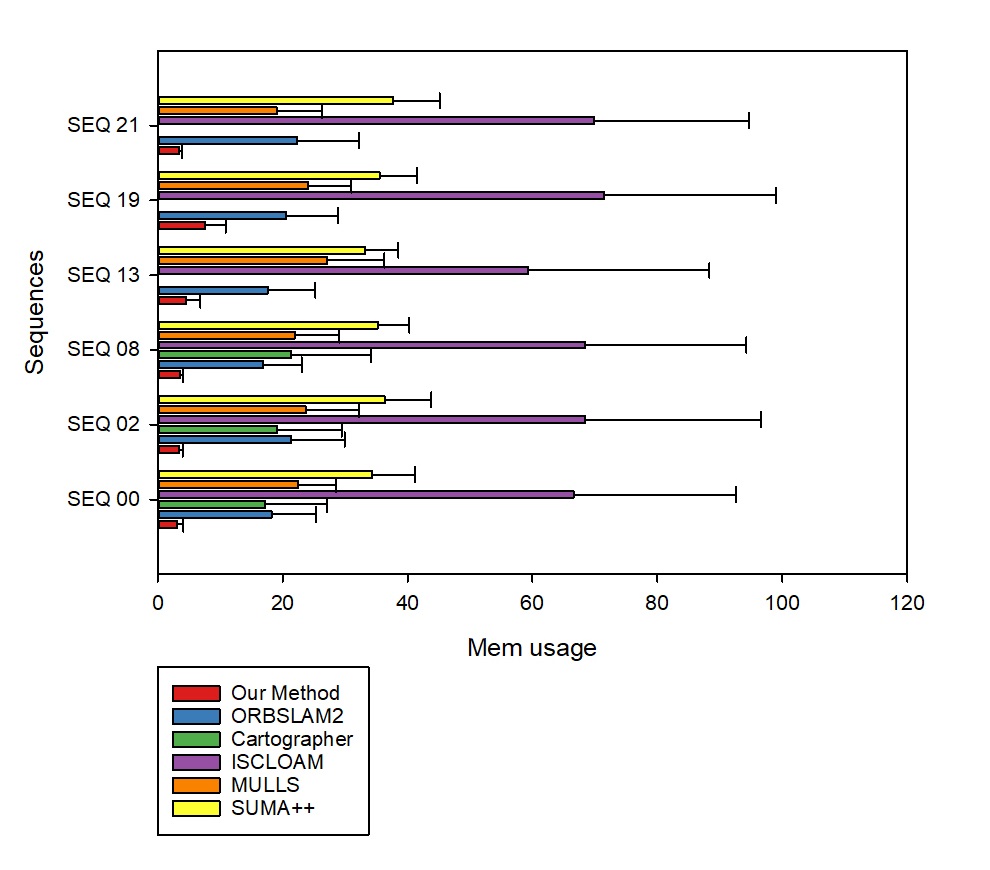}
	\caption{A plot of the memory requirements of the six methods for KITTI dataset sequences.}
	\label{fig:mem_us}
\end{figure}

Next, we examine the memory requirements of the online SLAM operation. Figure-\ref{fig:mem_us} shows the percentage memory usage of the six methods. This comparison shows the true impact of our local map representation and loop detection approach. We limit the memory usage by using local map representation and then keeping only recent local maps online. Next, the computational requirements are limited by the implementation of a BOW-based loop closure and re-localization. The memory usage for our system remains lower than 5\% for all the sequences. The remaining five systems report memory usage of up to 20\% and more, ISCLOAM has the highest values of around 60\%-80\%. Their system is a typical SLAM approach where new map points are added as the robot moves. All the frames are kept online for loop closure detection, which requires such high memory usage. SUMA++ reported lower CPU usage, but their system's memory requirements are much higher. ORBSLAM2 also uses the BOW approach using the DBOW2 library, but their computation requirements are much higher than our system. The main difference in performance is due to the size of the vocabulary. We use a small BOW database built by only using images of local maps limiting the memory usage. From these results, we have been able to prove the computational efficiency of our system.

\subsection{Indoor Dataset}

We have collected data inside our lab, with multiple runs of the area. Our lab presents a dynamic environment with constant changes in the scene. This kind of data is similar to the real-life operation of service robots that have to operate for a long time with people moving and other changes to the scene. In this section, we investigate the performance in two steps, i.e., the computational costs of the SLAM system during a long-term operation and the map quality. We use ORBSLAM2, ISCLOAM, and MULLS for evaluation with our system. We have not used SUMA++ in this section, as it only works well with the KITTI dataset.

\begin{figure}[hbt!]
	\centering
	\begin{subfigure}{.4\textwidth}
		\centering
		\includegraphics[scale=0.5]{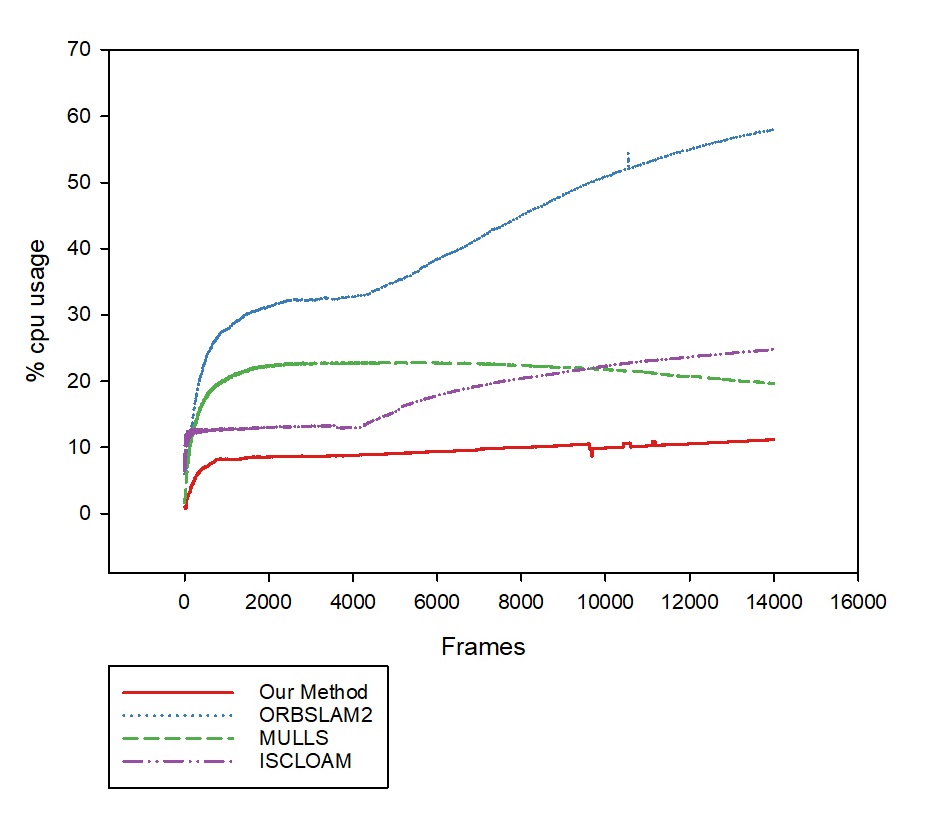}
		\caption{CPU usage}
	\end{subfigure}\qquad
	\begin{subfigure}{.4\textwidth}
		\centering
		\includegraphics[scale=0.5]{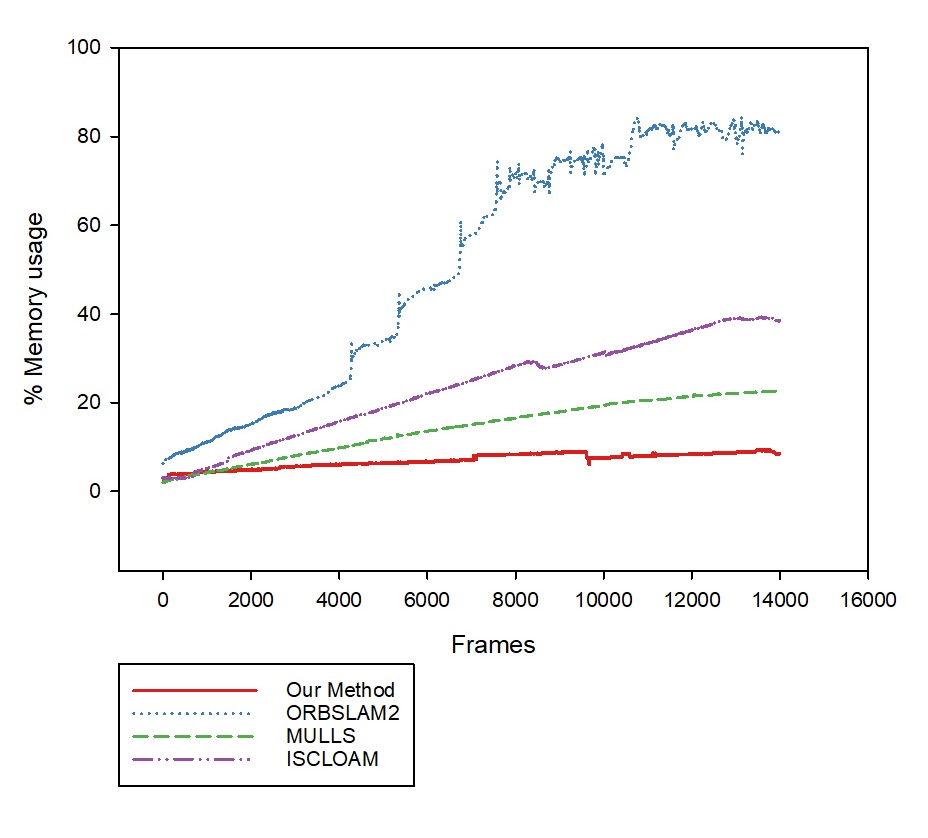}
		\caption{Memory Usage}
	\end{subfigure}
	\caption{Percentage cpu and memory consumption by our method, ORBSLAM2, ISCLOAM and MULLS for the indoor dataset}
	\label{lab_comp}
\end{figure}

One of the challenges for life-long SLAM is to limit the computational requirements of online operation. As the robot moves, new frames and map points keep on adding to the system. For most methods, this information is kept online for loop closure detection and global optimization. Figure-\ref{lab_comp} shows the percentage CPU and memory usage of the four systems used in this evaluation. First, we look at the performance of ORBSLAM2, as it reports the highest values of the four systems. The reasons are the vocabulary size for loop closure detection and relocalization, and the information kept online increases with the number of frames. 

ISCLOAM uses the approach to keep all the frames and map points online to search for loop closure. So, the computational requirements rise as the robot moves through the environment. We can also see from the results that as the number of frames grows, both CPU and memory requirements increase. MULLS reports different behavior in terms of CPU load. Its value is slightly higher than our system, but it almost remains constant. But the memory usage rises as the number of frames increase. Lastly, we can observe that plot for CPU and memory usage remains flat for our method. The CPU load stays less than 10 \%, and memory requirements are lower than 8\%. These results show the effect of the proposed strategy for long-term operation. In this paper, we have proposed an efficient mapping approach using local map rasterized images. We also ensure that only limited information is kept online and still produce accurate results.
Next, the loop closure detection method is fast and efficient. For indoor datasets with thousands of frames, loop closure thread only takes 1\% of memory usage. From these results, we have shown the efficacy of the proposed method to solve the computational complexity issues for long-term SLAM operation.

\section{Conclusion}

In this paper, we presented an approach of a light weight SLAM algorithm that can be applied to long-term operation. We divided the global map into a set of local map rasterized images. These images reduce the computational and memory requirements for mapping of SLAM system. A map management system keeps the online operation lightweight, even for longer operations. Using Bag of words approach, an efficient loop closure and re-localization method is also presented. We have provided a thorough evaluation of our method with experiments on KITTI dataset and an indoor dataset. For KITTI, our method reported recall and precision of above 90 percent and much lower CPU and memory usage. The indoor dataset was used to show the performance during longer operations. The computational requirements of our system remain constant for the indoor dataset.  

\bibliographystyle{IEEEtran}  

\bibliography{JIRS}

\end{document}